# Generative Adversarial Networks (GANs): An Overview of Theoretical Model, Evaluation Metrics, and Recent Developments

Pegah Salehi[†], Abdolah Chalechale[*], Maryam Taghizadeh[‡]

*Image Processing Research Lab, Department of Computer Engineering & Information Technology*
*Razi University, Kermanshah, IRAN*

[†]*pghsalehi@gmail.com*
[*]*chalechale@razi.ac.ir*
[‡]*taghizadehmail@gmail.com*

One of the most significant challenges in statistical signal processing and machine learning is how to obtain a generative model that can produce samples of large-scale data distribution, such as images and speeches. Generative Adversarial Network (GAN) is an effective method to address this problem. The GANs provide an appropriate way to learn deep representations without widespread use of labeled training data. This approach has attracted the attention of many researchers in computer vision since it can generate a large amount of data without precise modeling of the probability density function (PDF). In GANs, the generative model is estimated via a competitive process where the generator and discriminator networks are trained simultaneously. The generator learns to generate plausible data, and the discriminator learns to distinguish fake data created by the generator from real data samples. Given the rapid growth of GANs over the last few years and their application in various fields, it is necessary to investigate these networks accurately. In this paper, after introducing the main concepts and the theory of GAN, two new deep generative models are compared, the evaluation metrics utilized in the literature and challenges of GANs are also explained. Moreover, the most remarkable GAN architectures are categorized and discussed. Finally, the essential applications in computer vision are examined.

***Keywords:*** Deep learning, Deep generative models, Generative Adversarial Networks, Semi-supervised learning, Unsupervised learning

## 1 | Introduction

Recent several decades have witnessed a rapid expansion in artificial intelligence knowledge and its application in various sciences following an increase in the power of computational systems and the emergence of large datasets in different industries.

Machine learning[1], as one of the broad and extensively-used branches of artificial intelligence, is concerned with the adjustment and exploration of the procedures and algorithms based on which computers and systems develop their learning capabilities. Machine learning algorithms need to extract features from raw data. In previous methods, these features were manually provided and fed to the algorithm concerned, a time-consuming and incomplete task under certain circumstances. Representation learning or Feature learning[2] offers the system the ability to automatically discover the representations required for feature detection, classification, and other issues. In other words, representation learning transforms input data into meaningful outputs. Deep learning[3] is a kind of representation learning intended to model super-abstract concepts in the dataset according to a set of algorithms. This process is modeled using a deep graph consisting of several layers of linear and nonlinear transformations. Fig. 1 illustrates these definitions in the structure of the hierarchy.

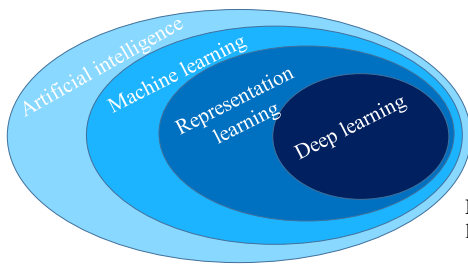

**Fig. 1.** Artificial intelligence, machine learning, representation learning, and deep learning at a glance.

---

[*] Corresponding author

Machine learning algorithms broadly separated into two main categories – supervised learning and unsupervised learning. Supervised learning needs a dataset with various features where each data should be labeled. These types of algorithms used to solve classification and regression problems. In contrast, unsupervised learning requires a dataset with more than one similar label. In this type of learning, the network is not told what pattern to look for, and there is no clear error metric. Some common examples of unsupervised learning include generative models, density estimation, clustering, noise generation, and noise elimination.

In supervised learning, manual management/collection of labeled data is costly and time-consuming; besides, automated data collection is also difficult and complicated. In deep learning, one of the vital tricks to solve the problem is the data augmentation method. Applying this method to the model increases the skill of the model, creates a regular effect, and reduces generalization error. Data augmentation is done by creating new and acceptable samples of the training dataset, including the application of operators, such as rotation, cropping, zooming, and other simple transformations on images. Nevertheless, only data with limited information can be obtained using this method. The state-of-the-art type of data augmentation is the generation of high-quality samples through generative models. Hence, considering the ability of generative networks to generate images on a large scale, it is expected that the severe shortage of labeled data will be substantially mitigated.

Generative models commonly work based on the Markov chain, maximum likelihood estimation (MLE), and approximate inference. Restricted Boltzmann Machine (RBM)[4] and its developed models such as Deep Belief Network (DBN)[5], and the Deep Boltzmann Machine (DBM)[6] are based on MLE. Generated samples by these methods compare the data distribution with the experimental distribution of the training data. These prototypes have several severe constraints and may not be well generalized.

Generative Adversarial Networks (GANs) were proposed as an idea for semi-supervised and unsupervised learning by Ian Goodfellow[7]. Yann LeCun, director of the IBC's research at Facebook, introduced adversarial training as the most interesting idea of the past ten years in machine learning[8]. Fig. 2 clearly shows the rapid growth in the number of published articles in the field of GANs in recent years. GANs have shown impressive improvements over previous generative methods, such as variational auto-encoders or restricted Boltzmann machines. Fig. 3 shows GANs progress over several consecutive years for face generation.

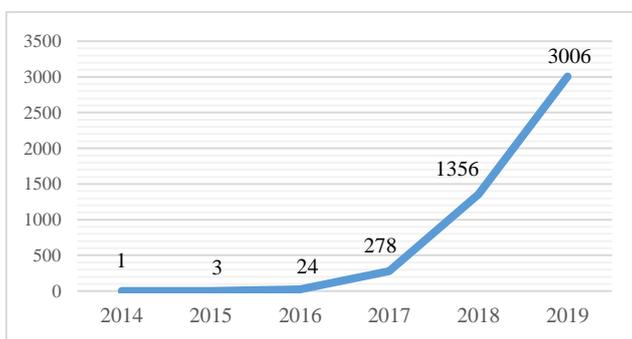

**Fig. 2.** Number of articles indexed by Scopus on GANs from 2014 to 2019. The chart from[9].

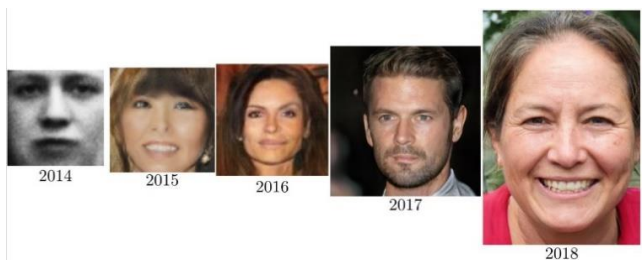

**Fig. 3.** 4.5 years of GAN progress on face generation. 2014[7], 2015[10], 2016[11], 2017[12], 2018[13].

So far, research has been conducted to review generative adversarial networks; In total has been dealt with the introduction of GANs, its applications in various fields such as computer vision[14], signal processing[15], image synthesis and editing[16], speech processing[17], how to combine the GAN with an autoencoder[18], introducing the most notable architectures of GAN[19], and investigating the relationship between GANs and parallel intelligence[20].

The main idea of GAN is inspired by a two-person zero-sum game where the profits (or loss) of a participant are precisely equal to the losses (or profits) of the other. The total gains of the participants minus the total losses will be zero. The GAN architecture consists of two networks that train together: i.e., the generator and the discriminator. The generator tries to learn the statistical distribution of real data to generate fake data that is indistinguishable from real-world data to mislead the discriminator into thinking of these as real inputs. In contrast, the discriminator is a classifier that discriminates whether a given content looks like real data from the dataset or like an artificially synthesized data. As both participants continuously optimize themselves to improve their capabilities and attempt to learn from their own weaknesses and take advantage of each other's weaknesses, the neural networks become stronger during the training process. The optimization process aims to establish a Nash equilibrium between the two participants. In economics and game theory, Nash equilibrium is a stable system state involving interaction between various participants. Under these circumstances, no participant can benefit simply by unilaterally changing the strategy without altering the strategy of the other participants, exactly what GAN is trying to do. Generator and discriminator reach a state where one cannot progress without changing the other.

Nowadays, GANs are widely used in various examples, such as text-to-image synthesis, image-to-image translation, and many potential medical applications. Fig. 4 shows the percentage of the total number of articles published until 2019 in different disciplines.

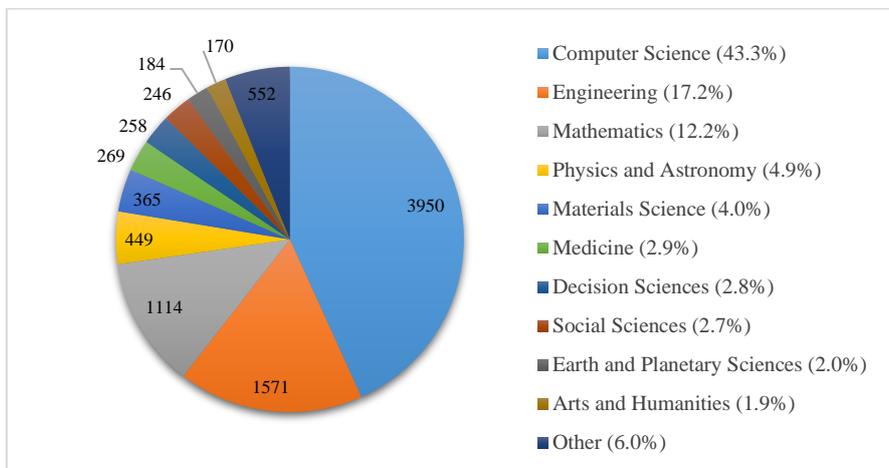

**Fig. 4.** Taxonomy of the number of articles indexed in Scopus based on different disciplines from 2014 to 2019. The cart from[9].

Given the importance of GAN and its application in various scientific fields, it is necessary to introduce it comprehensively, to investigate research carried out in this field, and to describe the challenges in this field. Therefore, this paper has addressed these issues. It is worth noting that a better understanding of GANs requires the perception of the concepts of deep learning. In the book[21], are introduced the basics of deep learning theory and the mathematical details. In another book[22], the common themes and concepts of deep learning are explained by coding in the Python Programming language.

This paper is structured as follows. In section II, the main concepts and the theory of GAN are clarified, then after comparing two recently introduced deep generative models, the evaluation metrics and challenges facing GAN are described. Section III lists the GAN architectures and addresses the most prominent and widely-used ones. Section IV describes some of the significant applications of GAN in the field of computer vision. Finally, Section V presents conclusions and new directions.

## 2 | Generative Adversarial Networks (GANs)

Generative adversarial networks are a potent class of neural networks that follow an intelligent approach to unsupervised learning. The GANs become able to generate samples very similar to the real data distribution through automated exploration underlying structure and learning of the existing rules and patterns of the real data [23].

The GAN framework naturally takes up a game-theoretic approach. GANs usually containing two neural networks to train and compete against each other: one generator and one discriminator. The reason for choosing the word "adversarial" in GAN is that these two networks are in constant conflict throughout the training process. These two networks can be likened to counterfeiter (generator) and police (discriminator). The generator attempts to create a form of money similar to real-world money by learning the latest tricks to deceive the police, i.e., the discriminator. Conversely, the police must continuously update their information to spot counterfeit money. The two networks are continually updating their knowledge and getting feedback on their successful changes. This struggle continues until the police fail to distinguish real data from fake data; this means that the counterfeiter is generating valid samples[24].

The architecture of GAN is illustrated in Fig. 5. $X_{data}$ and $G(z)$ are the real samples in the training dataset and fake samples synthesized by the generator $G$, respectively. Discriminator $D$ judges the probability that the input data is real or fake. In GAN, first, the generator takes noise vector $z$ (the random vector with uniform distribution or Gaussian distribution) of a fixed-length as input. Then, the generator synthesizes new data $G(z)$ from standard signal distributions $X_{data}$. To gain a better understanding of the problem, arguably, generating an image requires not an initial input of an image but a vector of random values. After training, the points of this multi-dimensional vector are matched with the points in the problem domain, resulting in a compressed representation of the data distribution. This vector space is known as a latent space or a vector space consisting of latent variables. Latent variables include important, yet unobservable variables directly for a domain[21]. Machine learning models can learn the statistical latent space of images, music, and stories and subsequently create a series of new artworks with specifications similar to those of real samples of this space[22].

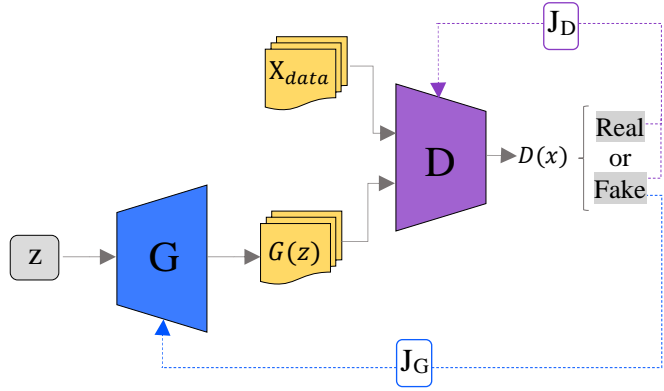

**Fig. 5.** The architecture of the GAN.

A discriminator acts as a binary classification and differentiates fake $G(z)$ samples from real $X_{data}$ samples. The discriminator is trained to maximize the likelihood of assigning the correct labels to real and fake data. In other words, if the input is made up of real $X_{data}$ data, the discriminator classifies it as real data and returns a numeric value close to 1. Otherwise, if the input is composed of data generated by the generator, the discriminator classifies it as fake data and returns a numeric value close to 0.

The generator and the discriminator can be neural networks, convolutional neural networks, recurrent neural networks, and autoencoders. Therefore, the discriminator requires the loss function $J_D$ and the generator requires the loss function $J_G$ to update the networks (Fig. 5). The generator updates its parameters only through the backpropagation signals of the fake output. By contrast, the discriminator receives more information and updates its weights using fake and real output.

GAN can be modeled as a two-player minimax game with simultaneous training of both generator and discriminator network. Minimax GAN Loss is regarded as an optimization strategy in two-player games whereby each player reduces their losses or increases the costs of the other player. In GAN, the generator and discriminator represent the two players, which in turn update their network weight. Minimax refers to minimizing the loss in the generator and maximizing the loss in the discriminator[25]. Put differently, the discriminator seeks to maximize the probability of assigning proper labels to the data. On the contrary, the generator seeks to generate a series of samples close to the real data distribution to minimize cross-entropy.

$$if \quad X = X_{data} \Rightarrow D(X) \to 1 \quad \Rightarrow \quad \max_{D} V(D.G) = E_{x \sim p_{data}(x)}[\log(D(x))] \tag{1}$$

$$if \quad X = G(Z) \Rightarrow \begin{cases} D(X) \to 0 \text{ ; for D} \Rightarrow \max_{D} V(D.G) = E_{z \sim p_z(z)}\left[\log\left(1 - D(G(z))\right)\right] \\ \\ D(X) \to 1 \text{ ; for G} \Rightarrow \min_{G} V(D.G) = E_{x \sim p_{data}(x)}[\log(D(x))] \end{cases} \tag{2}$$

$$\min_{G} \max_{D} V(D.G) = E_{x \sim p_{data}(x)}[\log(D(x))] + E_{z \sim p_z(z)}\left[\log\left(1 - D(G(z))\right)\right] \tag{3}$$

One reason that remains challenging for beginners is the topic of GAN loss functions. The GAN optimization strategy, as a minimax problem, is presented as Equation 3. For better understanding, Equation 3 is broken down into Equations 1 and 2. In which, $E$ is the mathematical expectation notation. $p_{data}$ stands for the distribution of real data, while $p_z$ is the random noise distribution.

According to Equation 1, if $X = X_{data}$ ($X$ is the input data to the discriminator), the discriminator is expected to display a numeric value close to 1 in the output. That is, $X$ is expected to distribute real data and maximize $V(G.D)$.

According to Equation 2, if $X = G(Z)$, there will be two different perspectives that address the first criterion of the problem from a discriminator perspective. The discriminator is expected to manage to detect that the generated sample is fake and to display a numeric value close to 0 in its output. $V(G.D)$ should also be maximized under these circumstances. It shows the second criterion of the problem from a generator perspective. Here, the ideal case for the generator is to be able to mislead the discriminator, i.e., a numeric value close to 1 is displayed in the output. In other words, the generator is trained to fool the discriminator by minimizing $V(G.D)$ and obtaining real data distribution. Finally, from a mathematical point of view, Equation 3 shows a 2-player minimax game with value function $V(G.D)$.

Fig. 6 illustrates several steps of the simultaneous training of generator and discriminator in a GANs as an example. In Fig. 6(a), GANs are trained by simultaneously updating the discriminative distribution (blue, dashed line) so that it distinguishes between samples from the real data distribution (black, dotted line) and generated data distribution (green, solid line). In Fig. 6(b), the discriminator trained to discriminate between real and fake data, and it easily does its task. In Fig. 6(c), the discriminator training process is stopped, and only the generator is trained to bring the fake data distribution closer to the real data distribution. These updates continue until the discriminator no longer distinguishes (Fig. 6(d)). It is worth noting that the process of training GANs is not as simple and straightforward as the process presented in Fig. 6 The fake data distribution is completely overlaid to the real data distribution under ideal conditions, while there are various challenges in practice.

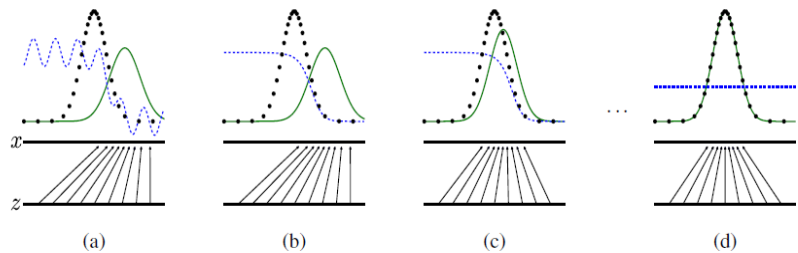

**Fig. 6.** An example of a GANs training process. Evolution of the generated data distribution (green) towards the real data distribution (black) and the decision boundary (blue). The figure from[7].

GANs are a group of networks with a very complex and challenging training process because both generator and discriminator networks are trained simultaneously in an adversarial manner. The whole basis of GANs is the equilibrium between the two networks. In other words, the nature of the optimization problem changes every time the parameters of one of the networks are updated, resulting in the establishment of a dynamic system. The technical challenge facing the training of two competing neural networks is their delayed convergence[25].

## 2.1 | A Comparison Between Two Deep Generative Models

The purpose of applying artificial intelligence is to provide machines with the ability to understand the complex world of humans. Inspired by this idea, researchers have proposed several generative models capable of describing the world around them statistically and probabilistically. In addition to the two popular generative models, i.e., RBM and the DBN, which make use of deep learning algorithms, two commonly used generative models were introduced in 2014, called Variational Autoencoder (VAE)[26] and GANs[7]. Both VAE and GAN methods attempt to learn the statistical distribution of real-world data, albeit with different teaching methods. Below these two generative models are briefly compared.

Almost everything is faced with uncertainty in the real world. Artificial Intelligence seeks to solve real-world problems; therefore, one of the challenges it has to tackle is addressing the uncertainties in real-world problems. The use of contingency models is one of the effective ways to overcome this challenge. VAE is essentially a probabilistic graphical model (PGM) that has its roots in Bayesian inference. It effectively and efficiently deduces under uncertainty using PGMs. The VAE purpose is latent modeling, and that is why it tries to model the probability distribution of latent data to obtain new samples. Among its applications: generating many kinds of complex data, including handwritten digits, faces, house numbers, CIFAR images, physical models of scenes, segmentation, and predicting the future from static images.

Compared to GAN, VAE employs a specific method for evaluating model quality. Furthermore, this method is more stable in training than GAN. However, it directly uses the mean squared error (MSE) to compute the latent loss function. It often generates blurry images compared to GAN because it is an extremely straightforward loss function applicable in a latent space. Nonetheless, GAN gradually improves the quality of the generated data using an adversarial training process and generates realistic and colorful pictures that a human can hardly distinguish it from real photographs.

## 2.2 | Evaluation Metrics

GAN is a practical deep learning approach for the development of generative models. Generally, deep learning models are trained until the convergence of the cost function. However, GAN exploits the balance between the generator and the discriminator for training. Thus, one of the problems with using adversarial networks to make a fair comparison is to evaluate the strengths and weaknesses of various models. To the best of the researchers' knowledge, no consensus has been reached so far on the estimation of relative or absolute quality and developed cost function training. Notwithstanding, a set of quantitative and qualitative methods are widely used categorized into three classes – manual evaluation, qualitative evaluation, and quantitative evaluation (see Fig. 7). In the following, a full explanation of each class is presented.

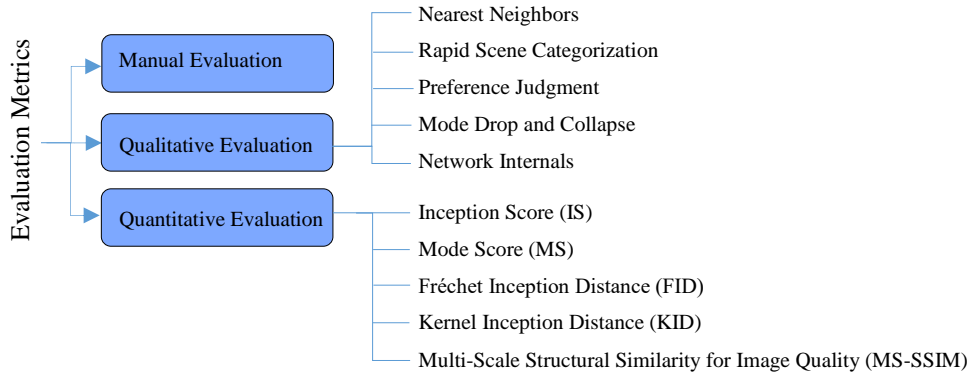

**Fig. 7.** Categorization of GAN evaluation metrics

1. Manual evaluation is a technique used to evaluate the quality and diversity of images generated. In this technique, the generated image and the target image are compared by the researcher himself/herself or by a person with related expertise. Visual inspection of samples by humans is one of the most common and intuitive GAN evaluation methods[27].

Like other deep learning models, a generative model is also trained in each epoch. The exact time to stop the training process and save the final model for subsequent use cannot be detected because there is no proper metric of model performance. Hence, the model can be saved regularly once every other epoch. Then, the saved model can be selected by manual inspection of generated images. This evaluation method is considered as a good starting point for beginners to get acquainted with the proposed architecture.

Being the most straightforward model evaluation method, manual evaluation involves numerous limitations. Evaluating the quality of images from a personal point of view is a relative and arbitrary issue; bias may be included in the comparison. Furthermore, it is expensive and time-consuming.

2. Qualitative evaluation is a series of non-numeric metrics, often involving comparative or subjective evaluation. Five qualitative methods for evaluating GAN models have been proposed[27]: a) Nearest Neighbors b) Preference Judgment c) Rapid Scene Categorization d) Mode Drop and Collapse e) Network Internal.

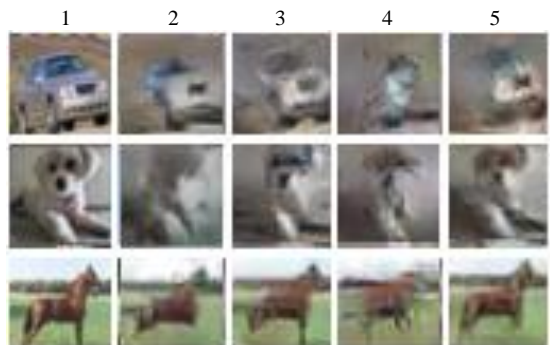

**Fig. 8.** Generated samples nearest to real images from CIFAR-10. In the first column shows real images, followed by the nearest image generated by DCGAN[10], ALI[28], Unrolled GAN[29], and VEEGAN[30], respectively.

The "nearest neighbors" method is one of the most well-known approaches in evaluating the performance of a generator model, which selects some samples of real images and one or more identical images for comparison (Fig. 8). Distance metrics such as Euclidean distance between pixel image information are often used to select the generated sample most similar to the real image. The nearest neighbors approach can help evaluate the degree to which the generated image is real.

The "preference judgment" method is one of the qualitative evaluation techniques, an extension of manual evaluation. In this type of experiment, individuals are asked to rate their generated images in terms of accuracy.

The "rapid scene categorization" method is similar to the former except that the images are shown to human judges for a split second, and they are asked to classify them into real or fake. The variance in judgment is reduced by averaging the scores among different judges. Being a complicated and time-consuming method, it can reduce costs by using a crowdsourcing platform such as Amazon Mechanical Turk (MTurk). Another disadvantage of this approach is the unstable performance of human judgment that can be improved over time.

One of the significant shortcomings in the development of GAN is the "mode drop and collapse". Mode drop occurs when the training process results in different outputs for similar inputs. Mode collapse, on the other hand, means that the diversity of the generated samples for different latent spaces is limited. In[30]–[32], several methods have been introduced to evaluate "mode drop" and "mode collapse."

Though the "network internals" inspection and visualization is a broad topic, it can be applied to find out which features in the latent layers are considered. The quality of internal representations can be evaluated by studying how the network is trained and understanding what it learns in the latent layers.

3. Numerical scores also are calculated to compare the quality of the images generated. In the following, some of the most widely used metrics for quantitative evaluation are discussed.

"Inception Score" (IS) metric[33] is an objective evaluation method to evaluate two features, i.e., the quality and the diversity of generated images. It employs the MTurk platform to evaluate a large number of generated images, indicating that IS performs well as subjective human evaluation. This metric has been introduced as an attempt to eliminate subjective human evaluation. IS uses the pre-trained inception v3 network [34], with its minimum value being 1.0. Higher IS values suggest the high quality of generated samples. IS is considered to be a useful and widely-used metric; however, when the generator reaches mode collapse, it may still display a good value.

An evaluation metric called "Mode Score" (MS) is introduced in[35] based on IS, which can concurrently reflect the diversity and visual quality of the generated samples. It has overcome the problem with IS, namely insensitivity to previous distributions of ground truth labels (i.e., disregarding the dataset)[27].

The "Fréchet Inception Distance" (FID)[36] is an improved IS metric that can be applied to detect inter-class mode dropping. In this method, the generated samples are embedded in the feature space provided by a particular layer of the inception network. The mean and the covariance between the generated samples and the real data are calculated, assuming that the generated samples follow a multidimensional Gaussian. The FID between the two Gaussians is then calculated to evaluate the quality of the generated samples (examples). Nevertheless, IS and FID cannot solve the overfitting problem well. In order to overcome the problem, the "Kernel Inception Distance" (KID) is offered in[37].

"Multi-Scale Structural Similarity for Image Quality" (MS-SSIM)[38] differs from the "Single-Scale Structural Similarity for Image Quality" (SS-SSIM)[39], which can be used to measure the similarity between two images. In this metric, the similarity between images is evaluated using Predicting Human Perception Similarity Judgment. In[40], [41], has been used this metric to determine the diversity of the data generated. Also, in[42], FID and IS were used as auxiliary evaluation metric with MS-SSIM to examine sample diversity.

In general, choosing an appropriate evaluation metric remains a complex issue. In[27], several measurements have been introduced as meta-metrics to guide researchers towards the selection of quantitative evaluation metrics. An appropriate evaluation metric should distinguish between the generated samples and real samples. Moreover, it should manage to detect mode collapse, mode drop, and overfitting. It is attempted to introduce more suitable techniques in the future to evaluate the quality of GANs.

## 2.3 | Challenges

Like any other technology, GANs also face several challenges. These problems are generally linked to the training process, including mode collapse and training process instability. Furthermore, the evaluation technique, image resolution, and ground truth are considered as other controversial domains.

One of the main issues of failure in GAN training is mode collapse. This refers to the state in which the generator starts generating similar images. In other words, the diversity of generated samples is limited to different latent spaces. One possible solution to increase data diversity is to use sample batch production instead of generating a sample. Another approach is to use multiple generators to obtain various samples. In[43], the generates combinatorial samples have been examined by different models to resolve the mode collapse. The objective function optimization can also be used to mitigate this challenge, similar to the WGAN[44] and unrolled GAN[29] models. Hence, how the diversity of generated samples should be increased is a crucial issue to be addressed in future work.

"Training process instability" is regarded as another challenge in this area, resulting in different outputs for similar inputs. Although batch normalization is considered as a solution to GAN instability, it is not sufficient to improve GAN performance with optimal stability. Numerous approaches have been suggested for more sustainable training[33], [36], [44]–[47]. Notwithstanding, several solutions should be proposed to train a more stable GAN and to converge on the Nash equilibrium. The next issue is GAN evaluation, a more complex issue than other generative models. In section 2, we review several currently extensively-used evaluation metrics. Providing appropriate, acceptable, and inclusive evaluation methods is one of the essential issues that need further study.

Another limitation of adversarial networks is the resolution of generated images. Currently, most GAN-based applications for image processing are limited to 256×256. When this network is applied to high-resolution images, some blurry images are usually created. Although some researchers use iterative coarse-to-fine methods to generate high-resolution images, they do not run fast. Chen and Koltun (2017) have introduced cascaded refinement networks to create a series of 2-megapixel images, a new perspective for high-resolution image production[48]. Collectively, it is possible to offer an appropriate approach to enhance the resolution with the flexibility of image size thanks to the excellent capabilities of the adversarial networks, and one area is still under investigation.

Using ground truth data for training is another common challenge; That is also known as a crucial problem in deep learning. These data play a vital role in the synthesis and editing of GAN-based images because real synthesized/edited images are not easily collected. The approaches proposed by CycleGAN[49] and Adversarial Inverse Graphics Networks (AIGNs)[50] use unpaired data for model training. These approaches can be regarded as a suitable solution to similar problems. Thus, this issue also requires further attention, investigation, and research into GAN-based applications.

## 3 | Types of GAN Models

Since the emergence of GAN, several GANs variants have been developed from the original GAN. That can be divided into two classes, developments based on architecture optimization, and developments based on objective function optimization, as shown in Fig. 9.

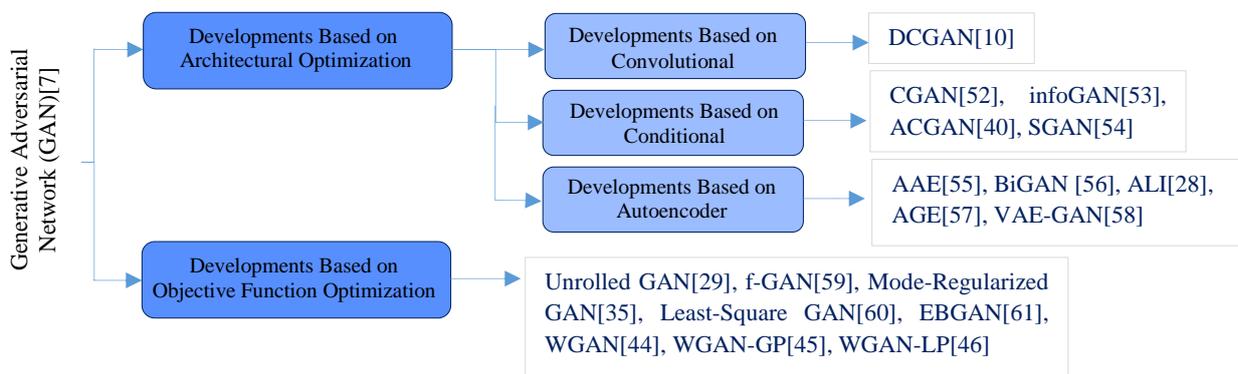

**Fig. 9.** Categorization of GAN models.

### 3.1 | Developments based on architecture optimization

As shown in Fig. 9, developments based on architecture optimization are divided into three classes; Convolutional, conditional, and Autoencoder. In the following, a full description of these classes is presented.

### 3.1.1 | Developments Based on convolutional

The Convolutional Neural Network (CNN)[51] is one of the most effective and widely-used learning models, with significant progress in the field of computer vision over the last few years. In an original GAN, the multilayer perceptron (MLP) is employed for generator and discriminator networks. The Deep Convolutional GAN (DCGAN)[10]architecture was proposed due to the unstable and challenging training of MLP and the better performance of CNN compared to MLP in feature extraction. In this architecture, CNN is used for the generator and discriminator network. Several changes were made to CNN so that it could be applied to the generator and discriminator network. These changes were obtained upon numerous trial and errors in architecture, configuration, and training plans. Table 1 shows the changes made to CNN to apply it to GAN. DCGAN can well generate high-resolution images, and it can also be said one of the most critical steps in designing and training sustainable GAN models. Most GAN models are based on this architecture.

In the DCGAN architecture, the generator has to capture random points in the latent space as input and generate an image. The proposed method realizes this goal by using a transposed convolutional layer that performs the inverse of the convolution operation (i.e., deconvolution). In other words, the stride of 2 will have an inverse effect, i.e., the upsample operation will be used instead of the downsample operation in the standard convolutional layer. In Fig. 10, the structure of the generator is shown.

Table 1. Changes applied to CNN architectures to use it in GAN.

|  | standard CNN | CNN (Generator) | CNN (Discriminator) |
| --- | --- | --- | --- |
| Dimensionality Reduction Layer | Pooling Layers | Fractional-Strided Convolutions | Strided Convolution |
| Batch Normalization | Not necessary | Necessary | Necessary |
| Activation Functions | Various activation functions | ReLU in all layers and Tanh in the last layer | leaky ReLU in all layers |
| Fully Connected Layer | Have | Does Not Have | Does Not Have |

A discriminator is a standard convolutional network that captures an image as input and displays a binary classification (real or fake) as output. In standard mode, deep convolutional networks utilize pooling layers to reduce input dimensionality and feature maps with the depth network. This is not recommended for DCGAN; instead, strided convolution is used for dimensionality reduction.

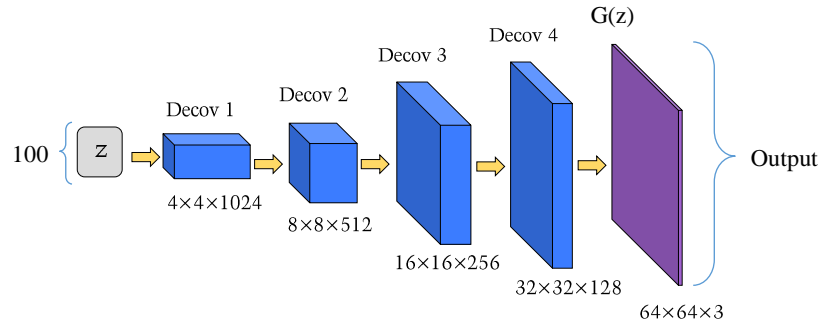

Fig. 10. The generator of deep convolutional GANs.

### 3.1.2 | Developments Based on Conditional

One of the limitations of the original GAN is that the generator randomly generates samples from the dataset. Indeed, there is no way to control the content of the generated image. The only existing solution is to understand the relationship between the latent space and the generated images, a complex issue with complicated mapping.

Therefore, conditional GAN (cGAN)[52] was introduced to overcome the random generation problem. This architecture is one of the various GANs that generates images conditionally by the generator. Some datasets contain additional information such as class labels, which had better be used to control generated images. Thus, image generation can be conditional on the class label. For example, the MNIST database of handwritten digits has a class label corresponding to its integer that can be used to generate specific numbers, such as 7.

According to Fig. 11(a), the condition variable $c$ helps the generator to generate a specific sample based on a condition. In other words, in a cGAN, the generator is trained with random noise $z$ with additional information $c$. Besides, the sample contains additional information $c$ fed to the discriminator to determine whether it is real or fake. Therefore, the loss function can be defined as,

$$\min_G \max_D V(D.G) = E_{x \sim p_{data}(x)}\left[\log D\big((x|c)\big)\right] + E_{z \sim p_z(z)}\left[\log\Big(1 - D\big(G(z|c)\big)\Big)\right] \tag{4}$$

Thanks to its immense popularity and influence, this architecture is widely used. Information GAN (infoGAN)[53] is developed from the cGAN architecture that makes the generation process more controlled. For example, in the MNIST database of handwritten digits, controls such as style, thickness, and type are used to generate the image of the handwritten digit. The cGAN architecture uses the label $c$ in the dataset, while infoGAN also extracts other latent features by the discriminator model and the probabilistic network $Q$. The image is fed to the discriminator as input, and realness or fakeness besides $Q(c|X)$ is displayed as output. $Q(c|X)$ is the probability distribution of $c$ conditional on image $X$ (Fig. 11 (b)). For example, the generated image of digit 3 is fed to the discriminator, and $Q$ may estimate (0.1, 0, 0, 0.8, …); the image will be "0" with a probability of 0.1 and "3" with a probability of 0.8.

The value of mutual information should be maximized to improve the relationship between $x$ and $c$. The generator of this architecture is similar to the cGAN architecture, except that the latent code $c$ is not known and must be discovered through the training process. The loss function is described as follows,

$$\min_G \max_D V(D.G) - \lambda I(c, G(z.c)) \tag{5}$$

$\lambda$ is a hyper-parameter to limit function $I(c, G(z.c))$. Mutual information makes latent codes $c$ more suitable for generated data.

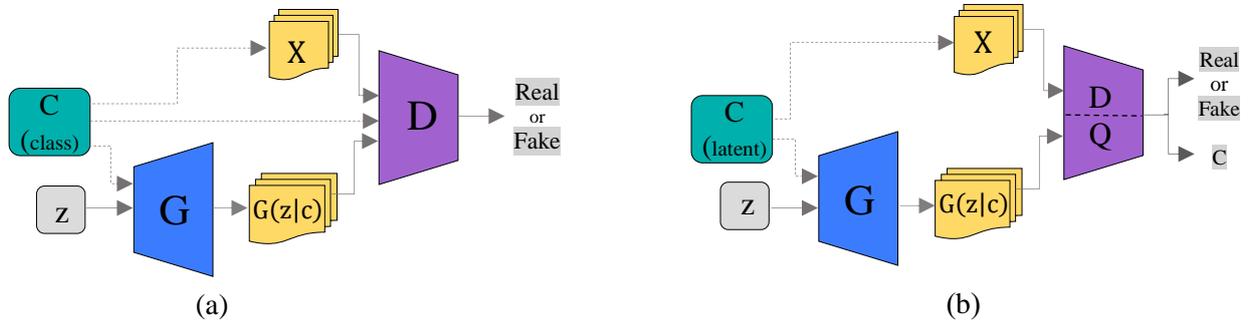

**Fig. 11.** The model of (a)CGANs; (b) InfoGAN

InfoGAN specifically explores visual concepts such as hairstyles, the presence or absence of glasses, and facial expressions. Although infoGAN is an unsupervised approach, experiments suggest that this architecture learns several interpretable representations comparable to representations learned by supervised methods.

Auxiliary Classifier GAN (AC-GAN)[40] is developed from cGAN. In this architecture, the class label $c$ is not inserted into the discriminator. It uses another classifier to predict the probability of class labels $c$ besides the probability of the degree to which the image is real. In this method, the training process becomes more stable, and the model can generate higher quality images in larger sizes.

Semi-Supervised GAN (SGAN)[54] is developed from the GAN architecture that simultaneously trains supervised discriminator, unsupervised discriminator, and generator. One of the main goals of this architecture is to improve the performance of adversarial networks for semi-supervised learning. The discriminator is updated by predicting N+1 classes, where N is the number of datasets and classes added to match with the output of the generator.

### 3.1.3 | Developments Based on Autoencoders

Autoencoder neural networks are a type of deep neural networks used for feature extraction and reconstruction operations. This network consists of two parts, the encoder $z = f(x)$, and the decoder $\hat{x} = g(z)$. The encoder converts $x$ into a latent layer $z$ through an input dimensionality reduction process. At the same time, the decoder reconstructs the input $x$ to output $\hat{x}$ by receiving the code from the latent layer $z$. The autoencoder architecture is considered an unsupervised model because labels are not required during the training process. In the past few years, autoencoder networks have been used in deep generative models. One of the disadvantages of the autoencoder network is that the latent layer generated by the encoder is not distributed evenly over the specified space, resulting in a large number of gaps in the distribution.

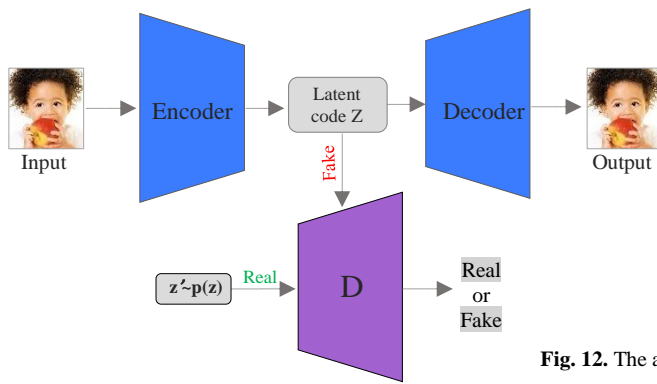

**Fig. 12.** The architecture of Adversarial Autoencoder (AAE)

Hence, the adversarial autoencoder (AAE)[55], a combination of the adversarial network with the autoencoder, was presented. In this approach, the previous arbitrary distribution is imposed on the latent layer distribution obtained by the encoder to ensure that no gaps exist so that the decoder can reconstruct meaningful samples from each part of it. The AAE architecture is illustrated in Fig. 12. In this architecture, the latent code $z$ represents fake information, and $z'$ is represented by the specified distribution $p(z)$, with both inputs acting as the discriminator. Upon completion of the training process, the encoder can learn the expected distribution, and the decoder, on the other hand, can generate the samples reconstructed by the required distribution.

Some models add an encoder to the GAN[28], [56], [57]. The generator of these models can learn the features of the latent space and obtain semantic variations in the data distribution; however, it cannot learn to map the distribution of the data sample to the latent space. To address this problem, bidirectional GAN (BiGAN)[56] was introduced, not only capable of generating valid inferences but also guaranteeing the quality of generated samples. The BiGAN architecture is illustrated in Fig. 13(a) In this architecture, an encoder is added to the model in addition to the discriminator and generator. The encoder uses the inverse mapping of data generated by GANs. The discriminator input for the generated data consists of a tuple containing the generated data $G(z)$ and the corresponding latent code $z$. Another discriminator input for real samples from the dataset is a tuple containing samples $X$ and $E(x)$ obtained by the inverse mapping of $X$ by the encoder. In this method, the encoder can be used as a feature extractor for the discriminator. Similar to the BiGAN architecture, an adversarially learned inference (ALI)[28] was offered. This architecture employs an encoder to obtain the distribution of the latent feature. These two approaches can simultaneously do generator and encoder learning.

In addition to the approaches that used a combination of autoencoder/adversarial networks, the Adversarial Generator-Encoder (AGE) Network architecture[57] is proposed. In the adversarial architecture, the generator and the encoder compete with each other without requiring a discriminator. Fig. 13(b) illustrated the AGE architecture in which R represents the reconstruction loss function. According to the structure of this model, the generator tries to minimize the divergence between the latent distribution $z$ and the distribution of the generated data. On the other hand, the encoder seeks to maximize the divergence between $z$ and $E(G(z))$ as well as to minimize the divergence of real data $X$. Moreover, the reconstruction loss has been used to prevent mode collapse. In Section 2-1 of this paper, we briefly compared deep generating models VAE and GAN. In[58], the benefits of GAN and VAE are combined, in that the VAE decoder is combined with the alternative GAN generator and the GAN loss function with the VAE objective function. This can reduce the severity of their difficulty in generating blurry images by preserving VAEs' capability in learning latent code distribution.

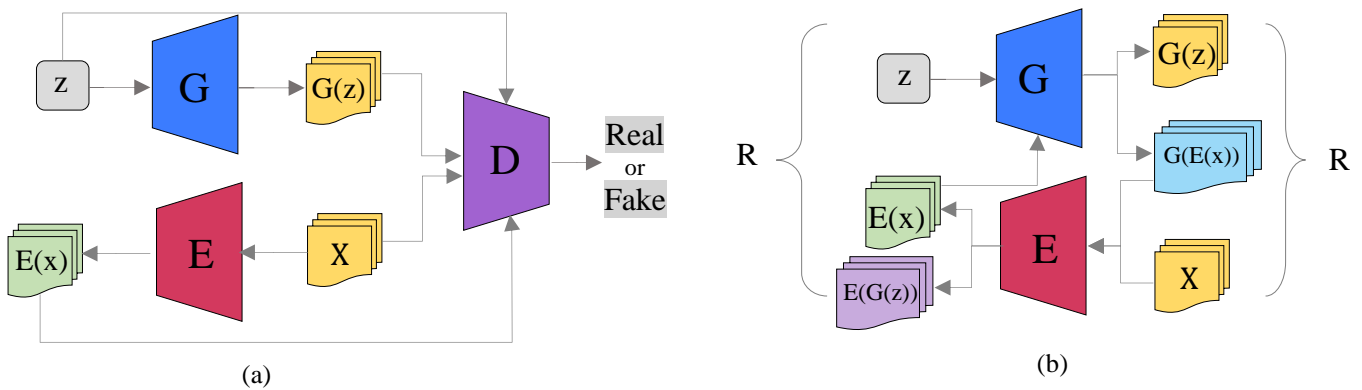

**Fig. 13.** The model of (a) BiGAN and ALI; (b) Adversarial Generator-Encoder Network(AGE)

## 3.2 | Developments based on objective function optimization

Several methods have been suggested to optimize the objective function to increase the stability of GANs, some of which will be briefly described below. In[29], unrolled GAN is offered to regulate the training processes of GANs. This method uses the gradient-based loss to strengthen the generator; however, original GANs attempt to minimize the Jensen-Shannon (JS) divergence to minimize the generator loss. Research[59] suggests that any divergence can be employed in the architecture of GANs. In[59]–[61], different divergences have been used to construct the objective function to enhance the stability of GANs.

Other regularizations are also used to improve the stability of GANs. In work[35], two regulators have been applied to learn further sustainability. Divergence tends to remain constant if there is no overlap (or a negligible overlap) between the distributed generated data and the real data. It causes the gradient to become zero and to lead to the vanishing gradient problem. In order to solve this problem, Wasserstein GAN (WGAN) has been proposed[44]. This method uses Earth-Mover (EM) or Wasserstein-1 distance estimation instead of JS divergence. It has also been shown theoretically that the EM distance generates better gradient behaviors compared to other distance metrics in distributed learning. This approach provided a weight clipping method to apply Lipschitz constraints. A new loss metric was also found to address the unstable learning process problem.

Due to the use of weight clipping in discriminator, WGAN may generate undesirable results or not converge. Hence,

WGAN with Gradient Penalty (WGAN-GP) was suggested to apply the Lipschitz constraint[45]. This method performs better than standard WGAN because it can train all kinds of GAN architectures in advance in a more stable way, almost without any meta-parameter setting. Additionally, in[46], a new penalty was proposed for applying the Lipschitz restriction as to the WGAN Lipschitz Penalty (WGAN-LP). This method effectively improves network training sustainability.

## 4 | Applications of GAN in Computer Vision

GAN has performed tremendously well than traditional methods in many fields of computer vision. According to attributes of adversarial mechanism and continuous self-improvement, GAN has a high-ability at learning features from existing distributions and capturing suitable visual features. In this section, we will introduce some leading-edge applications of GANs, including image super-resolution, image-to-image translation, face image synthesis, and image inpainting.

### 4.1 | Image super-resolution

Image super-resolution (SR) has been widely used in satellite, medical, and military images, and more. Deep learning techniques helped solve this problem by predicting the high-frequency details lost in low-resolution images. Super-Resolution GAN (SRGAN)[62] has been introduced to improve image resolution. In this method, a low-resolution image is received, and a high-resolution image is generated at a 4x scale.

The texture information generated by SRGAN is not real enough and is accompanied by noise. Enhanced Super-Resolution GAN (ESRGAN)[63] was introduced to solve this problem. In this network architecture, the adversarial loss and perceptual loss have been improved. Moreover, a new network called Residual-in-Residual Dense Block (RRDB) based on relativistic GANs[64] has been introduced. As shown in Fig. 14, ESRGAN performs better than SRGAN.

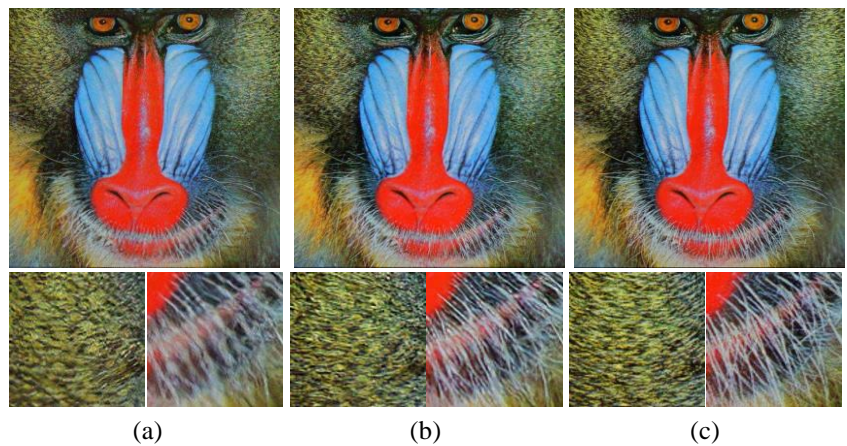

(a)          (b)          (c)

**Fig. 14.** Image super-resolution generated by different GANs. (a) SRGAN[62]; (b) ESRGAN[63]; (c) Ground truth.

### 4.2 | Image-to-image translation

As automatic language translation is considered a fundamental problem in machine learning, the image-to-image translation problem is similarly raised. The image-to-image translation is an unsupervised controlled technique to return an image from one representation to another.

The translation of the image from one domain to another is a challenging problem, often requiring a specific model and loss functions for the dataset. Classical methods employ classification or regression per pixel. In these methods, each of the output pixels is predicted with respect to the input image independent of the earlier pixels, leading to the loss of a large portion of the semantic content of the image. Ideally, however, a general approach capable of using a model and the loss function is needed for several image-to-image translation problems.

The Pixels-to-Pixels (Pix2Pix)[65] method is introduced using cGAN architecture to solve this problem. With its high potential, this architecture can generate real, high-resolution images in various image-to-image translation applications. It also allows for the creation of large images (e.g. 256×256) compared to older GAN models. Fig. 15 presents the performance of Pix2Pix. In the Pix2Pix architecture, the generator is inspired by the U-Net[66] and the discriminator by PatchGAN[67]. Both networks utilized in this architecture are deep convolutional neural networks. In PatchGAN, the classification is conducted in one step for all of the images. Instead, each image is divided into n×n sections at first. Then, whether the image is real or fake, is predicted separately for each patch. Ultimately, the final classification is performed by averaging all the responses. That is to say, PatchGAN penalizes the structure at the scale of patches. The pix2pix has shown its capability in a variety of applications, such as in stain normalization of histopathology images, it was used as stain-to-stain translation (STST)[68].

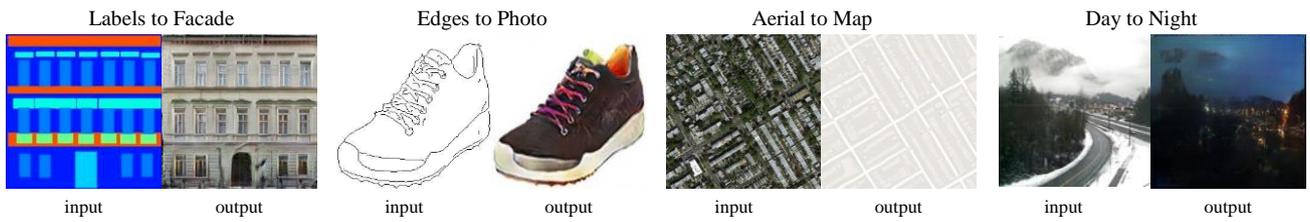

**Fig. 15.** Image translation generated by the pix2pix model[65].

Further studies in this field have proposed pix2pixHD[69] to promote the resolution of generated samples. In this method, a new adversarial loss function is used to generate images with a resolution of 2048×1024. Pix2pix requires to train the paired dataset, which is one of its limitations. That is, a dataset must be constructed from the input images before translation and the output images from the same images after translation. However, such image pairs do not exist in many cases. The unpaired image-to-image translation method in cycle-consistent GAN (cycleGAN)[49] can be employed to overcome this problem. This method uses cycle consistency loss, which seeks to preserve the original image after a translation and inverse translation cycle. This cycle does not need to pair images for training (see Fig. 16).

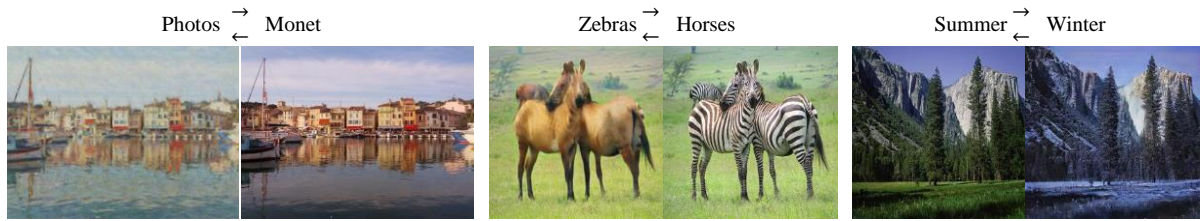

**Fig. 16.** Image translation generated by the CycleGAN model[49]. This model automatically translates an image from one into the other and vice versa.

### 4.3 | Face image synthesis

Face synthesis is widely used, including in face recognition. Although numerous data-based deep learning approaches have been proposed to do so, this area remains challenging. Since human vision is sensitive to facial deformities and deformations, generating real face images is no easy task. GANs have demonstrated to be capable of generating high-quality face images with fine texture.

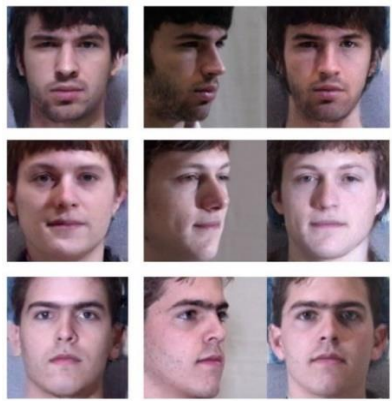

**Fig. 17.** Frontal face image synthesis through profile images using the TP-GAN method[70].

Two-Pathway GAN (TP-GAN)[70] can use a profile image to generate high-resolution frontal face images (see Fig. 17). This technique can consider local and global information like human beings. The face image generated by this method well preserves the characteristics of an individual's identity. It can also process multiple images in different modes and lighting. It has a two-pathway architecture. A global generator is trained to generate global features and a local generator to generate details around the face markings (marked points).

Furthermore, Self-Attention GAN (SAGAN)[71] combines self-attention block with GAN for image synthesis to solve long-range dependency problem. Thus, the discriminator is confident that it can determine the dependency between two distant features. In this approach, the improvement of the quality of the synthesized image is of greater importance.

Based on SAGAN, the BigGAN method[72] is proposed to increase the diversity and accuracy of generated samples by increasing the batch size and using a truncation trick. In the traditional approach, for the latent distribution $z$, $z$ is fed to the generator as input. Nevertheless, in BigGAN, $z$ is embedded in multiple layers of the generator to affect the resolution characteristics and different levels. As shown in Fig. 18, the generated samples are realistic.

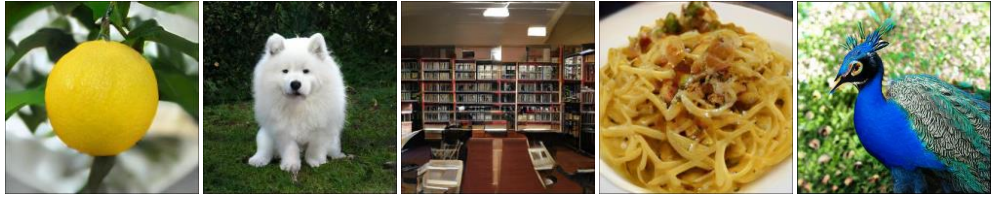

**Fig. 18.** Image synthesis generated by the BigGAN model[72].

The Disentangled Representation Learning GAN (DRGAN) method[73] was introduced for face image synthesis in the new state. The generator uses an encoder-decoder architecture that learns separate representations for face images, encoder output, and decoder input. The discriminator contains two parts, i.e., identity classification and state classification. The results of the experiments show that DRGAN outperforms the existing face recognition techniques in a steady state.

The face frontalization GAN (FF-GAN) architecture[74] uses a 3D morphable model (3DMM)[75] in the GAN structure. 3DMM provides geometry and appearance for face images. Likewise, 3DMM representations are small in volume. Fast FF-GAN convergence and high-resolution full-faced images are of high quality.

### 4.4 | Image inpainting

Image Inpainting seeks to reconstruct the lost parts of an image so that observers fail to spot the reconstructed areas. This method is often used to remove unwanted objects from an image or restore damaged parts in old images. In traditional techniques, the holes in the image were filled by duplicating the pixels of the original image or a library of images. Deep learning-based approaches have yielded promising results for restoring large areas lost in an image. These methods can create acceptable image structures and textures.

Some of these techniques have been suggested using convolutional networks, with poor performance in filling gaps with the correct features. Hence, generative models were developed to find the correct features known in the training process. The first image restoration method based on GANs is presented as a context encoder[76]. This method is trained based on the encoder-decoder architecture to infer arbitrary missing large regions in images based on image semantics. Nonetheless, in this method, a fully connected layer cannot store accurate spatial information. Context encoder sometimes creates blurry textures in proportion to areas around the hole. In[77] then combined the idea of "style translation" with the context encoder and proposed a new approach to restore high-resolution images. However, this model is not powerful enough to fill the missing area with complex structures. In Fig. 19, a sample result of this method and Context Encoder is shown.

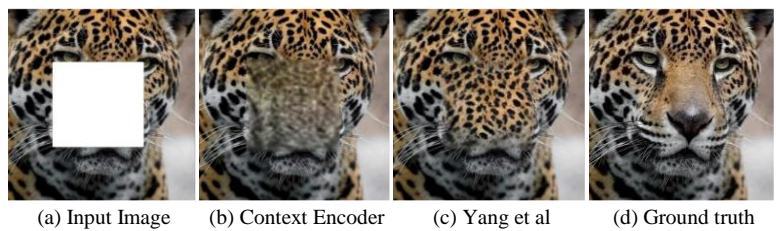

(a) Input Image    (b) Context Encoder    (c) Yang et al    (d) Ground truth

**Fig. 19.** Comparison between Context Encoder[76] and Yang et al.[77]

Likewise, in[78], the researchers used DCGAN to restore the image, which can generate lost parts of the image successfully. Nonetheless, there is still a blurry state at the hole border. In[79] has been proposed a GAN-based approach to image restoration compatible with global and local environments. The input is an image with an additional binary mask to display the missing hole. The output of a restored image has the same resolution. The generator employs the encoder-decoder architecture and extended convolutional layers instead of standard convolutional layers to support a larger spatial[80]. There are two discriminators, a global discriminator that captures the whole image as input and a local discriminator that covers a small region with its hole as input. The two discriminator networks ensure that the resulting image is compatible, on both "global" and "local" scales. This results in a natural restored image for high-resolution images with arbitrary holes.

## 5 | Conclusion and New Directions

In recent years, the generative adversarial networks (GANs) have been introduced and exploited as one of the widely-used deep learning algorithms and become a very popular architecture for generating highly realistic content. This architecture tries to generate data with similar characteristics as the input training data, which has caught the attention of many researchers thanks to its resistance to over-fitting in solving computer vision problems.

As deep neural networks require much data to train on, if data provided is not sufficient, they have poor performance. GANs can overcome this problem by generating novel and realistic data, without using tricks like data augmentation. A wide range of valuable research and practical applications are actively pursued in this field. Undoubtedly, with an improvement in network architectures and algorithms in the future, GANs will be expected to produce high-quality images, music files, movies, and texts, which are very difficult for humans to build. The development of GANs for applications such as text, natural language processing (NLP), and information retrieval (IR) undoubtedly yield significant results. The paper reviewed the main concepts and the theory of GAN, new models in this area, and also applicable evaluation metrics. Moreover, influential architectures and computer-vision based applications are examined.

Noteworthy, examining the link between GANs and reinforcement learning (RL) has been a growing research path over the last few years. Therefore, from a different angle point of view, the GANs are also regarded as one of the most promising recent developments for unsupervised learning. In some practical applications, appending a certain number of labels can significantly increase its production capacity. Therefore, how the GAN and semi-supervised learning are combined is one of the significant areas for future research.